\pdfoutput=1

\documentclass[11pt]{article}

\usepackage[preprint]{acl}
\usepackage{multirow}

\usepackage{times}
\usepackage{latexsym}

\usepackage{float}
\usepackage{microtype}
\usepackage{inconsolata}
\usepackage{graphicx}
\usepackage{soul}
\usepackage{booktabs}

\usepackage[T1]{fontenc}

\usepackage[utf8]{inputenc}

\usepackage{microtype}

\usepackage{inconsolata}

\usepackage{enumitem}

\usepackage{pifont}
\usepackage{xcolor}
\newcommand{\cmark}{\textcolor{green}{\ding{51}}} 
\newcommand{\xmark}{\textcolor{red}{\ding{55}}}   

\usepackage{array}
\newcolumntype{L}[1]{>{\raggedright\arraybackslash}p{#1}}
\usepackage[dvipsnames]{xcolor}

\title{Detecting Pipeline Failures through Fine-Grained Analysis of Web Agents}

\author{Daniel Röder, Akhil Juneja, Roland Roller, Sven Schmeier \\
German Research Center for Artificial Intelligence (DFKI) \\
\texttt{\{daniel.roeder, akhil.juneja, roland.roller, sven.schmeier\}@dfki.de}
}

\begin{document}
\maketitle
\begin{abstract}

Web agents powered by large language models (LLMs) can autonomously perform complex, multistep tasks in dynamic web environments. However, current evaluations mostly focus on the overall success while overlooking intermediate errors. This limits insight into failure modes and hinders systematic improvement. This work analyzes existing benchmarks and highlights the lack of fine-grained diagnostic tools. To address this gap, we propose a modular evaluation framework that decomposes agent pipelines into interpretable stages for detailed error analysis. Using the SeeAct framework and the Mind2Web dataset as a case study, we show how this approach reveals actionable weaknesses missed by standard metrics - paving the way for more robust and generalizable web agents.

\end{abstract}

\section{Introduction}

AI agents powered by large language models (LLMs) are increasingly deployed in real-world applications that require complex, multistep decision-making, such as coding assistance \cite{qiao2023taskweaver}, question answering \cite{liu2023gradually}, automated fact verification \cite{sun2023towards, XIONG2025104241} and web navigation \cite{ yin2024agent, he2024webvoyager, zheng2024gpt, zheng2025skillweaver}. These systems typically decompose tasks into modular pipelines, allowing structured reasoning across several intermediate steps. However, evaluation methods predominantly focus on end-to-end task success, offering limited visibility into intermediate reasoning and decision-making processes. This coarse-grained perspective obscures how and why agents fail - hindering systematic debugging, error diagnosis, and safe deployment in real-world scenarios where failures can lead to inefficiencies, degraded user experiences, or unintended behavior.

\begin{figure}
    \centering
    \includegraphics[width=\linewidth]{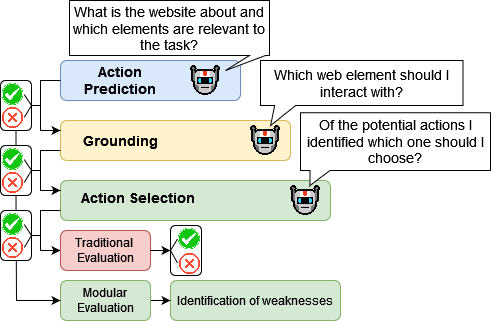}
    \caption{Modular evaluation of the SeeAct agent reveals performance across distinct pipeline stages.}
    \label{fig:modular_evaluations}
\end{figure}

Web navigation presents a particularly challenging setting for LLM-based agents, requiring multimodal reasoning over structured HTML, natural language, and visual elements. In this context, agents must interpret dynamic and often ambiguous information while executing a sequence of interdependent actions. Small errors in early stages - such as misinterpreting context or selecting the wrong subgoal - can propagate through the pipeline and lead to final task failure. Yet, most existing benchmarks and studies, such as those based on the Mind2Web dataset \cite{zheng2024gpt, deng2024mind2web}, evaluate only final task outcomes. This limits our understanding of where errors occur and how agent decisions break down in practice.

To address this gap, we propose a \textit{modular evaluation} that decomposes web agent pipelines into interpretable stages, including subgoal planning, information grounding, and action selection, and enables fine-grained error analysis across each step. Our goal is not only to expose failure modes more systematically, but also to motivate a shift toward diagnostic evaluation practices for LLM-based agents.

As a case study, we apply our modular evaluation to SeeAct \cite{zheng2024gpt}, a multimodal web agent that uses vision-language models (VLMs) to perceive and act within web interfaces. Our contributions are as follows:

\begin{itemize}[noitemsep,topsep=0pt,leftmargin=*]
    \item Introduction of a \textbf{a conceptual modular evaluation} that captures reasoning quality at each stage of the agent pipeline, providing a comprehensive error analysis.
    \item \textbf{Extending of the SeeAct architecture} by enhancing input representations and improving the heuristic action selection.
    \item \textbf{Augmentation of the Mind2Web evaluation protocol} by introducing alternative valid action annotations, addressing limitations of rigid single-ground-truth assumptions and better reflecting real-world flexibility.
\end{itemize}

Our experiments show that modular evaluation reveals systematic challenges such as context fragmentation, grounding errors, and ambiguity in HTML-based interfaces - issues often missed by standard metrics. By enabling step-wise diagnosis, our approach supports debugging, robustness testing, and system improvement - key capabilities for advancing reliable and generalizable LLM-based web agents. We release\footnote{\href{https://anonymous.4open.science/r/WebAgent-08F1/}{Anonymous Github Repo}} our evaluation toolkit and SeeAct reimplementation to facilitate further research in this direction.

\section{Related Work \& Background}

\subsection{Web Agents}

Web agents are systems designed to execute action sequences on web interfaces based on natural language instructions \cite{mazumder-riva-2021-flin, xu-etal-2021-grounding}. Traditional agents have struggled with diverse layouts and the complexity of web structures. Recent advances in large language models (LLMs) have enabled agents to infer contextually appropriate actions from natural language inputs, greatly expanding their versatility across tasks \cite{furuta2023multimodal, gur-etal-2023-understanding, sodhi2023heap, wang2024survey}.

A persistent challenge lies in effectively interpreting HTML content, which often lacks semantic clarity or task-specific grounding, making accurate action selection difficult \cite{deng2024mind2web, gur2023real, kim2024language, sridhar2023hierarchical}. To address this, several systems incorporate vision-language models (VLMs) that combine visual perception with language understanding—improving both generalization and task success rates. \citet{zheng2024gpt} introduced a multimodal agent that uses screenshots of web pages for visual understanding and acting on the web. \citet{hong2024cogagent} extended this idea with a generalist agent capable of reasoning across diverse visual domains. \citet{shahbandeh2024naviqate} further integrate multi-modal inputs to enhance contextual understanding in their functionality-guided navigation agent. Several other systems have also leveraged VLMs to boost generalization \cite{he-etal-2024-webvoyager, zhang-etal-2025-litewebagent}

\begin{table}[h!]
\centering
\small
\renewcommand{\arraystretch}{1.3}
\small
\begin{tabular}{@{} l c c l @{}}
\toprule
\centering \textbf{Web Agent} & \textbf{Planning \&} &  \textbf{Evaluation} \\
& \textbf{Grounding}  & \textbf{Granularity} \\
\midrule
\citet{he-etal-2024-webvoyager}  & \xmark & End-to-End \\
\citet{shahbandeh2024naviqate}  & \cmark & End-to-End \\
\citet{10.1145/3637528.3671620} & \cmark & End-to-End\\
\citet{iong-etal-2024-openwebagent}  & \cmark & End-to-End \\
\citet{zheng2024gpt} & \cmark & End-to-End \\
Ours & \cmark & Fine-Grained\\
\bottomrule
\end{tabular}
\caption{Comparison of evaluation granularity.}
\label{tab:agent_comparison}
\end{table}

\subsection{Evaluation of Web Agents}

Despite the increasing complexity of web agents, most evaluation protocols remain coarse-grained, focusing solely on the final, end-to-end task success, instead of taking also intermediate steps into account. \citet{zhou2023webarena}, for instance, evaluate agents based on final task correctness, and \citet{li2024websuite} introduce a taxonomy of web actions to categorize failure types. Other approaches such as \citet{pan2024webcanvas}, \citet{muhlbacher2024towards} and \citet{xu-etal-2025-crab} propose multi-level or composite metrics (e.g., \textit{Success}, \textit{Partial Success}), yet they still operate at the task level and do not expose failure propagation across intermediate steps.

These evaluations provide only limited insight into why agents fail, particularly in multistep settings in which reasoning, grounding and execution are interacting closely with each other. Without visibility into intermediate stages, debugging becomes difficult, and iterative improvement remains guesswork. In contrast, our work introduces a modular evaluation framework that decomposes the agent pipeline into interpretable stages, enabling fine-grained performance tracking at the level of planning and grounding. This not only supports targeted debugging but also opens the door to better understanding of failure sources and their frequency across tasks.

Table~\ref{tab:agent_comparison} compares recent web agents along key dimensions relevant to evaluation: whether they implement explicit planning and grounding, whether they support modular evaluation, and the granularity of their evaluation protocols. As shown, all prior work uses end-to-end evaluation, with no mechanism for inspecting or isolating failures within the pipeline. Our improved SeeAct agent is the only system to support modular evaluation with stage-level granularity, while also achieving a higher success rate than the original SeeAct baseline. This comparison highlights the broader need for diagnostic tools that go beyond task-level metrics to enable deeper analysis of web agent behavior.

\section{Methodology}

We propose a modular evaluation for LLM-based web agents that enables fine-grained diagnosis of reasoning and grounding failures. Rather than assessing only the final task outcome, our approach decomposes an agent pipeline into interpretable stages and evaluates each step independently. This helps to localize error sources, understand failure propagation, and guide system improvement.

\begin{figure*}
    \centering
    \includegraphics[width=\textwidth]{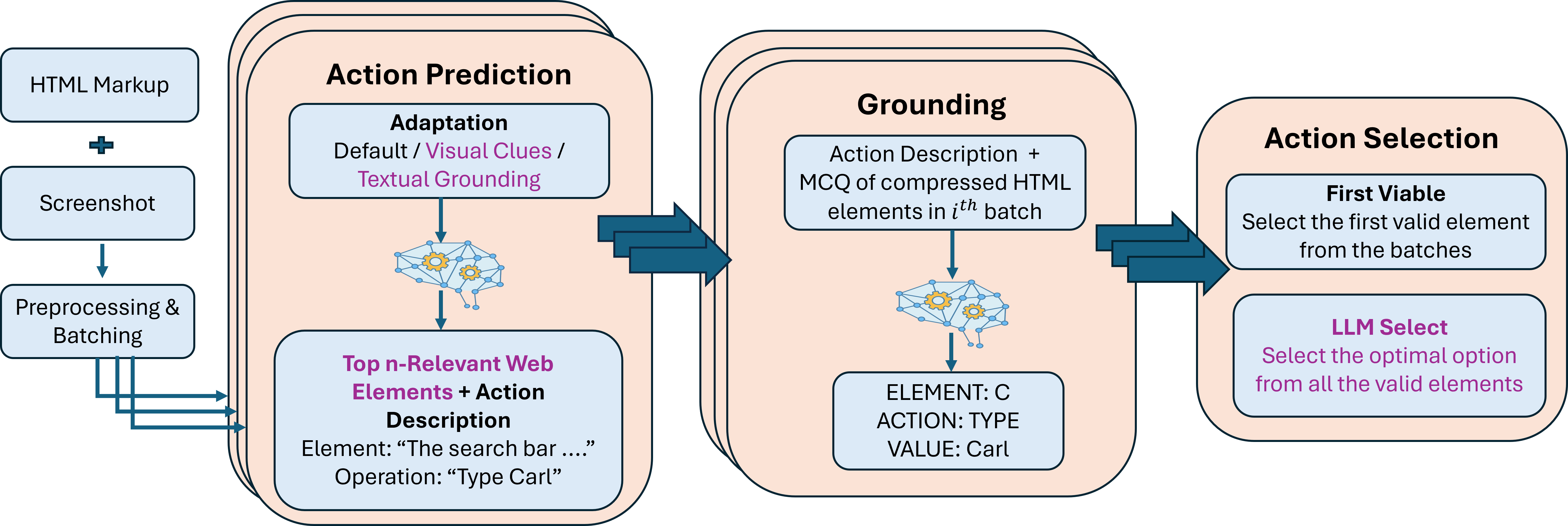}
    \caption{Overview of the adapted SeeAct pipeline. The process consists of three main stages: \textbf{Action Prediction} to generate an abstract action description, \textbf{Grounding} to map it to a specific HTML element, and \textbf{Action Selection} to choose the final command from multiple parallel batch outputs. Our adaptations, such as different \textbf{\textcolor{Purple}{Prompt Templates}} (e.g., Textual Grounding), \textbf{\textcolor{Purple}{Intermediate Reasoning}} and the final \textbf{\textcolor{Purple}{LLM Select strategy}}, are shown within this modular structure.}
    \label{fig:seeact_adaptation2}
\end{figure*}

\subsection{A Modular Evaluation Framework}
\label{subsec:mod_eval_framework}

Our framework evaluates web agents by decomposing their complex behavior into distinct, interpretable stages (Figure \ref{fig:modular_evaluations}). This modular approach allows for fine-grained diagnosis, as small errors in early stages can otherwise cascade and lead to task failure in ways that are obscured by end-to-end metrics. Each stage is evaluated with tailored metrics designed to reflect reasoning quality and alignment with the ground-truth action.

\paragraph{Stage 1: Action Prediction (Planning)}
In this initial stage, the agent observes its environment (e.g., HTML, screenshots) and, based on the task instruction, performs reasoning to identify relevant elements and decide on the next abstract action. We assess this stage with two metrics:
\begin{itemize}[noitemsep,topsep=0pt,leftmargin=*]
    \item \textbf{Relevant Element Accuracy (RE Acc.):} Measures the effectiveness of candidate generation. It is the percentage of instances where the ground-truth element is successfully included within the set of candidate elements presented to the agent.
    \item \textbf{Action Prediction Accuracy (AP Acc.):} Measures the core planning decision. Given a candidate set known to contain the ground-truth element, this is the percentage of times the agent's abstract prediction correctly identifies that element as the target for its next action.
\end{itemize}

\paragraph{Stage 2: Grounding.}
Here, the agent's abstract intent from the planning stage is translated into a concrete, executable action triplet (\textit{Element, Action, Value}). We evaluate this with a single, strict metric:
\begin{itemize}[noitemsep,topsep=0pt,leftmargin=*]
    \item \textbf{Grounding Accuracy:} The percentage of instances where the fully-grounded action triplet produced by this stage exactly matches the ground-truth triplet. For evaluations involving parallel batches, we only assess the batch that contains the ground-truth action in its candidate set.
\end{itemize}

\paragraph{Stage 3: Action Selection.}
This final stage resolves the outputs from the agents that have batch processing. Since each webpage section is processed as an independent batch, multiple viable, grounded actions are often generated. The task is to select the single correct action from this aggregated set of candidates. We assess two distinct strategies:
\begin{itemize}[noitemsep,topsep=0pt,leftmargin=*]
    \item \textbf{First Viable:} A simple heuristic baseline that measures the accuracy if we simply choose the first valid (non-None) candidate from the aggregated list.
    \item \textbf{LLM Select:} Our proposed strategy measures the accuracy of an LLM prompted to evaluate all available viable candidates and choose the single best one.
\end{itemize}

This modular structure supports plug-in evaluation on any agent pipeline that outputs intermediate predictions. We demonstrate its utility on the SeeAct agent and the Mind2Web benchmark.

\subsection{Case Study: SeeAct + Mind2Web}
\label{subsec:seeact_case}

To exemplify our framework, we apply it to the SeeAct web agent \cite{zheng2024gpt} and evaluate performance on the multimodal Mind2Web dataset \cite{deng2024mind2web}. Below, we briefly introduce both components and describe the adaptations we made to support modular evaluation.

\subsubsection{Mind2Web Dataset}
Mind2Web is a benchmark of 2,000 multistep tasks designed for web-based decision making (e.g., “Rent the cheapest SUV starting today”). Each task contains HTML structure, full-page screenshots, and ground-truth user actions. Action include CLICK, TYPE, SELECT, and most tasks require executing multiple sequential steps. One disadvantage of the dataset is, that it accepts (includes) only one single ground-truth path, even though multiple solutions are possible to solve a web-task. To support more flexible and meaningful evaluation, we augment Mind2Web with alternative valid actions (see Section~\ref{subsubsec:add_mined_data}).

\subsubsection{SeeAct Pipeline}
SeeAct is a multimodal agent that uses vision-language models (VLMs) to interpret HTML and screenshots. Given a task instruction and the current web page, it outputs an action triplet \textit{(Element, Action, Value)}. The original pipeline consists of following stages:

\begin{itemize}[noitemsep, topsep=0pt, leftmargin=*]
    \item \textbf{Preprocessing}: The top-50 task-relevant HTML elements are selected using a fine-tuned ranking model \cite{deng2024mind2web} and split into several spatial batches for parallel processing.

    \item \textbf{Planning}:  A VLM analyzes each batch and predicts the best next action as a natural language description.
    
    \item \textbf{Grounding}: For each batch, a LLM maps the natural language action description to one of the structured HTML candidates, producing a grounded action triplet. If no suitable match is present, then it selects None.
    
    \item \textbf{Action Selection}: A simple heuristic selects the first non-“None” grounded action from across all the parallel batches as the final output.
    
\end{itemize}

\subsection{Adaptations for Fine-Grained Analysis}
\label{subsec:seeact_adaptations}

To better support modular evaluation and address limitations in the original SeeAct pipeline, we introduce several key improvements. We illustrate our full adapted pipeline in Figure~\ref{fig:seeact_adaptation2}.

\subsubsection{Input Modifications}
\label{subsubsec:input-modifi}
A key weakness in the original framework is a potential mismatch of candidate elements between the Planning and Grounding stages. To eliminate this, we modified the pipeline to ensure that both stages operate on the identical set of candidate elements. We then created two adaptations to guide the agent's reasoning:

\begin{itemize}[noitemsep, topsep=0pt, leftmargin=*]
    \item \textbf{Textual Grounding (TG):} The top-k HTML elements are provided to the agent as a textual list with unique letter identifiers, forcing the VLM to operate within a constrained, known candidate set.
    \item \textbf{Visual Clues (VC):} Red bounding boxes highlight the top-k HTML elements directly on the screenshot, constraining the agent's visual focus.
\end{itemize}

The \textbf{Default (Def)} setting serves as our baseline and is defined as the configuration where neither the TG nor VC adaptations are applied.

\subsubsection{Enhancing Reasoning and Selection}
In addition to the input modifications, we implemented two foundational improvements that are applied across all experiments:

\begin{itemize}[noitemsep, topsep=0pt, leftmargin=*]
    \item \textbf{Intermediate Reasoning:} We prompt the VLM to first identify and explain up to five relevant elements before it selects a final action. This provides a chain-of-thought-like mechanism for more detailed analysis.
    \item \textbf{LLM-based Action Selection:} We replace the original heuristic ("first viable") selection strategy with a more robust LLM-based selector. Given all grounded candidates from the parallel batches, this module prompts an LLM to compare them and choose the single most plausible action.
\end{itemize}

\subsection{Augmented Evaluation with Flexible Ground Truth}
\label{subsubsec:add_mined_data}

To overcome the limitations of the original Mind2Web labels with only one valid solution, we identify alternative valid actions within a task, focusing on interchangeable steps. For example, when applying two successive filters (for instance to buy black shoes in size 10) “size” or “color” the order does not play any role to finish the task. 

We mine such alternatives by checking whether action-relevant HTML elements appear in multiple steps of the same task, using minimal CSS selectors. We treat these as valid substitutes when order does not affect task correctness. Evaluation results are reported with and without these alternatives, providing a more realistic assessment of agent performance.

\begin{table*}[ht!]
\centering
\small
\begin{tabular}{llcccccc}
\toprule
 Pipeline Stage:& &\multicolumn{2}{c}{Action Prediction} & & \multicolumn{2}{c}{Action Selection} \\
\cmidrule(lr){3-4} \cmidrule(lr){6-7}
 Adaption & Model & RE Acc. & AP Acc. & Grounding & First Viable & LLM Select \\
\midrule

\multirow{6}{*}{\centering Def}
    & Gemini-1.5-flash & 61.45 & 56.59 & 52.13 & 28.57 & 37.76 \\
    & Gemini-1.5-pro & 62.27 & 56.18 & 52.94 & 35.71 & 38.57 \\
    & Claude 3.5 Sonnet & 63.89 & 56.58 & 49.27 & 24.08 & 36.53 \\
    & InternVL2-Llama3-76B-AWQ & 61.45 & 54.16 & 49.90 & 17.96& 26.53 \\
    & GPT-4o & \textbf{74.44} & \textbf{70.17} & \textbf{62.87} & \textbf{48.78} & \textbf{54.29} \\
    & GPT-4o-mini & 61.25 & 57.18 & 52.13 & 28.78 & 33.88 \\

\bottomrule
\end{tabular}
\caption{Modular evaluation results for the six models under the standard Default (Def) adaptation, measuring accuracy at each stage of the agent pipeline. This data establishes a performance baseline.}
\label{tab:pipeline_analysis_def_adap}
\end{table*}

\begin{table*}[ht!]
\centering
\small
\begin{tabular}{llcccccc}
\toprule
 Pipeline Stage:& &\multicolumn{2}{c}{Action Prediction} & & \multicolumn{2}{c}{Action Selection} \\
\cmidrule(lr){3-4} \cmidrule(lr){6-7}
Adaptation & Model & RE Acc. & AP Acc. & Grounding & First Viable & LLM Select \\
\midrule

\multirow{3}{*}{VC}
    & Gemini-1.5-flash & 62.89 & 56.59 & 47.86 & 22.24 & 31.22 \\
    & Gemini-1.5-pro  & 62.48 & 57.80 & 48.47 & 22.65 & 31.02 \\
    & Claude 3.5 Sonnet  & 63.90 & 58.62 & 52.12 & 21.22 & 37.76 \\
    & InternVL2-Llama3-76B-AWQ & 68.96 & 55.39 & 51.12 & 18.98 & 30.41 \\
    & GPT-4o  & \textbf{78.29} & \textbf{75.04} & \textbf{67.54} & \textbf{44.08} & \textbf{53.27} \\
    & GPT-4o-mini & 69.57 & 60.63 & 50.91 & 25.51 & 31.84 \\ 
\midrule
\multirow{3}{*}{TG}
    & Gemini-1.5-flash  & 80.9 & 62.01 (67.54) & 60.26 & 21.22 & 37.35 \\
    & Gemini-1.5-pro & 79.72 & 66.74 (70.37) & 63.90 & 24.69 & 42.24 \\
    & Claude 3.5 Sonnet  & 76.07 & 63.7 (66.0) & 60.45 & 23.06 & 37.14 \\
    & InternVL2-Llama3-76B-AWQ & 82.36 & 64.31 (72.42) & 61.87 & 21.22 & 34.49 \\
    & GPT-4o & \textbf{83.79} & \textbf{72.83 (79.7)} & \textbf{69.59} & \textbf{31.84} & \textbf{51.02} \\
    & GPT-4o-mini & 73.22 & 60.47 (67.35) & 57.43 & 18.57 & 31.22 \\

\bottomrule
\end{tabular}
\caption{Modular evaluation results for six models across two adaptations Visual Clues (VC) \& Textual Grounding (TG), measuring accuracy at each stage of the agent pipeline. As TG employs a separate evaluation pipeline for AP Acc that removes ambiguity, we also present results of the default pipeline (in parentheses) for comparison.}
\label{tab:pipeline_analysis_all_model}
\end{table*}

\section{Experimental Setup}

We evaluate our approach using a combination of state-of-the-art open-source and proprietary language models, including \textbf{Gemini, Claude, InternVL2-LLaMA3} and \textbf{GPT-4} variants. All experiments are conducted on a 90-task subset (490 actions) of the multimodal Mind2Web dataset, using three test-set splits - website, domain and task as used by \citet{zheng2024gpt}. Detailed model configurations are provided in Appendix~\ref{appendix: model_specifications}.

For optimal performance, the VC and Def adaptations are evaluated using four webpage sections (batches), while TG is evaluated using five. Preliminary results for decision-making are provided in Appendix \ref{appendix_selecting_sections}.

Evaluating natural language outputs from the Def and VC adaptations in the planning stage requires matching free-form text descriptions to specific HTML elements. To perform this robustly, we employ a fine-tuned neural classifier. Specifically, we train a \textbf{BGE-Small-en-v1.5} encoder \cite{bge_embedding} using the SetFit framework \cite{tunstall2022efficient} on 800 labeled pairs of HTML-text matches. For the TG adaptation, evaluation is a direct string comparison of the selected identifier (e.g., "C") against the ground truth. Further evaluation details are available in Appendix~\ref{appendix:evaluating_rel_elems}.

\section{Results \& Analysis}

We present the results of our modular evaluation, consolidated in Table~\ref{tab:pipeline_analysis_def_adap} and Table~\ref{tab:pipeline_analysis_all_model}. We first establish the end-to-end performance of the agents to provide a "black box" view. We then leverage our fine-grained framework to diagnose the primary sources of failure within the pipeline, demonstrating the critical insights that a modular approach provides over a single metric.

\subsection{Illusion of a single metric: Pinpointing System-Wide Bottlenecks}

From a traditional end-to-end perspective (`First Viable' in Table~\ref{tab:pipeline_analysis_def_adap}), GPT-4o is the top-performing model with 48.78\% accuracy. However, this single metric masks the true nature of agent failures. Our modular analysis reveals two primary bottlenecks that systematically limit performance.

\begin{figure}[h]
    \centering
    \includegraphics[width=\linewidth]{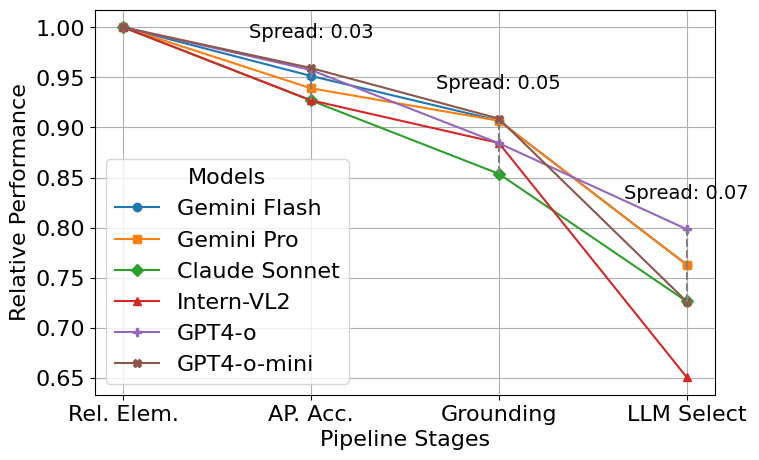}
    \caption{Relative performance decline across pipeline stages for the Def adaptation. Normalizing the initial score reveals a similar drop-off pattern across models, pointing to a shared bottleneck.}
    \label{fig:relative_perf}
\end{figure}

The \textbf{first major bottleneck is the initial `Action Prediction' stage (planning)}. Even the best model, GPT-4o, achieves only 70.17\% accuracy here, meaning nearly 30\% of tasks fail due to flawed initial reasoning before later stages are even attempted. The \textbf{second bottleneck is the final `Action Selection' stage}, where performance further drops. 
More critically, our analysis of relative performance decline (Figure~\ref{fig:relative_perf}) shows that despite varying absolute scores, all models exhibit a remarkably similar pattern of error propagation across the pipeline. This suggests that the bottlenecks are not model-specific but are systemic challenges inherent to the web navigation task itself, a key insight only visible through modular evaluation.

\subsection{Analysis of Adaptations}

To understand the nature of the bottlenecks identified in our pipeline, we now analyze how the TG and VC adaptations influence agent behavior. 

\textbf{The Textual Grounding (TG)} adaptation, which reframes element selection as a multiple-choice question (MCQ), consistently yields the highest Relevant Element and Grounding accuracy. \textbf{This confirms that a structured format is highly effective for isolated element identification}. However, this strength reveals a critical trade-off.

The core issue stems from a combination of the binding nature of TG and the agent's lack of global context. Unlike the Def setting where the agent can predict an action on an element not in the candidate set (often leading to a "None" action later), TG forces the agent to choose from the provided list. As each webpage section (batch) is processed in parallel without awareness of others, the agent is prone to selecting a "best-fit" action within nearly every section, even when no action is required. This results in a significantly higher number of viable but unnecessary candidate actions (avg. 3.91 per five sections). \textbf{This abundance of options ultimately overwhelms the final Action Selection stage}, explaining why TG's initial gains in identification do not consistently translate to higher final accuracy.

\textbf{The Visual Clues (VC)} adaptation, which provides bounding boxes on the screenshot, yielded highly model-dependent results. While it boosted performance for visually capable models like GPT-4o, it offered negligible benefit—or even harmed performance—for others. This suggests a performance threshold for many models, the cognitive overhead of parsing cluttered visual information and interpreting potentially occluded bounding boxes outweighs the directive benefit of the clues. \textbf{This highlights the current limitations in robust, general-purpose visual grounding, where the "help" provided by visual aids can instead become a source of noise}. Details on bounding box occlusion analysis are in Appendix \ref{appendix:bbox_occlusion}.

\subsection{Analysis of the Final Selection Stage}

The final selection stage, where the agent must commit to a single action from a set of grounded candidates, represents the last major hurdle. Across all configurations, \textbf{our `LLM Select' strategy consistently outperforms the simpler `First Viable' heuristic}, confirming the value of sophisticated reasoning at this final step. However, this stage remains a significant bottleneck.

Our analysis reveals that \textbf{performance is directly impacted by the number of viable options presented to the selector}. Additional information on viable actions and selection performance provided in Appendix \ref{appendix: num_viable_actions}. TG (avg. 3.91) and VC (avg. 2.93) adaptations resulted in an increased number of viable actions compared to the Def (avg. 2.5). As the number of choices increases, selection accuracy across models drops sharply, from an average of 73.1\% with two options, down to 56.0\% with four. This highlights a critical challenge: the agent's task is not just to identify a correct action, but to disambiguate it from other plausible alternatives.

\begin{figure}[h]
    \centering
    \includegraphics[width=0.9\linewidth]{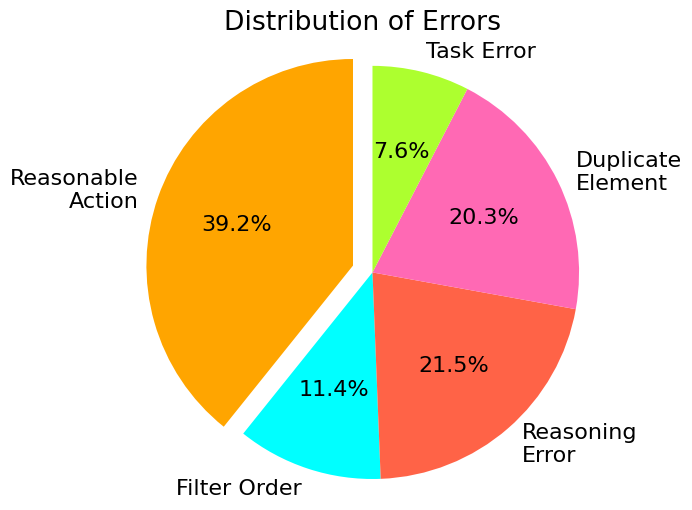}
    \caption{Manual classification of errors for a Gemini-1.5-Pro.}
    \label{fig:distribution of errors}
\end{figure}

This problem of ambiguity is exacerbated by two factors. First, certain adaptations naturally produce a larger set of viable candidates, increasing the difficulty of the final selection. Second, and more importantly, many of these alternatives are only "incorrect" because of the benchmark's rigid single-ground-truth assumption. Our manual error analysis (Figure~\ref{fig:distribution of errors}) confirms this, showing that over 50\% of errors are actually Reasonable Actions and Filter Orders.

\begin{figure}[h]
    \centering
    \includegraphics[width=\linewidth]{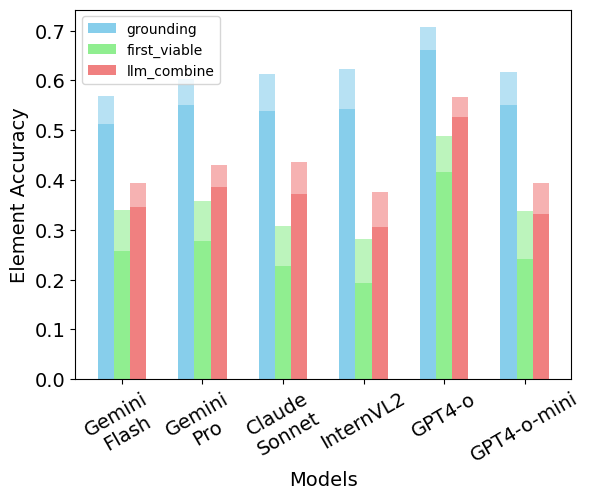}
    \caption{Added performance through additionally mined candidates (Light Shaded)}
    \label{fig:viable_cand_boost}
\end{figure}

To quantify the impact of this benchmark rigidity, we augmented the ground truth with a small set of these reasonable mined candidates. As shown in Figure~\ref{fig:viable_cand_boost}, this addition improved performance by up to 8.3\%, \textbf{directly demonstrating that a significant portion of failure at this stage is attributable to the benchmark's inflexibility}. While our `LLM Select' strategy handles this ambiguity better than a simple heuristic, these findings underscore the need for more dynamic and flexible evaluation protocols that better reflect real-world web interaction.

\subsection{Qualitative Error Analysis}

To better understand the nature of the failures identified by our modular framework, we manually classified errors from a Gemini-1.5-Pro run on Test-Website split (168 samples) into five categories. The distribution is shown in Figure~\ref{fig:distribution of errors}.

\begin{itemize}[noitemsep, topsep=0pt, leftmargin=*]
    
    \item \textbf{Reasonable Actions (39.2\%):} This category includes semantically correct but non-matching actions, such as clicking on the Nike brand page instead of the "Shoes" category when the task is to buy Nike shoes. Another example is choosing to use the search bar rather than applying filters. While these actions are valid and align with the overall task intent, they deviate from the recorded ground-truth sequence.

    \item \textbf{Alternative Filter Usage and Order (11.4\%):} These errors occur when filters are applied in a different order—for example, filtering by color first and then size, or vice versa. We address this issue of filter order variation with our extended ground truth candidate set (Section \ref{subsubsec:add_mined_data}).

    \item \textbf{Reasoning Errors (21.5\%):} These are genuine failures of logic, where the agent misunderstands the task or the webpage content, such as selecting a clearly incorrect item. This represents the core challenge for improving the models' planning capabilities.

    \item \textbf{Duplicate Element Errors (20.3\%):} These errors happen when the agent can't tell apart similar-looking elements, because the compressed HTML doesn't provide enough detail. The agent either selects a semantically similar element or fails due to insufficiently descriptive compressed HTML. This highlights a key weakness in the grounding stage. An example of ambiguous HTML scenarios in Appendix \ref{appendix:ambiguity} . 
    
    \item \textbf{Task Errors (7.6\%):} These small percentage of errors were attributed to suspected issues in the Mind2Web labeling process where actions don’t match the task or issues with reference screenshots.

\end{itemize}

This qualitative analysis reinforces our main quantitative findings that the two largest sources of failure stem not from trivial errors, but from the agent's difficulty with \textbf{benchmark ambiguity} (Reasonable Actions) and \textbf{poor grounding/selection signals} (Duplicate Elements).

\section{Discussion}

Our modular evaluation has uncovered several key insights with broader implications for the design and evaluation of web agents.

\paragraph{Pinpointing the Core Reasoning Failures}
A key implication of our work is that end-to-end metrics mask the true nature of agent failure. By deconstructing the pipeline, our modular analysis reveals that the agent's journey is bookended by two stages of intense difficulty: initial \textbf{Planning} and final \textbf{Action Selection}. While errors do occur at every step, these two high-level reasoning tasks represent the primary bottlenecks where the majority of performance is lost. This finding shifts the focus of future research away from uniform pipeline improvements and towards targeted advancements in the agent's core decision-making faculties: forming an effective initial plan and disambiguating the best final action from a set of plausible alternatives.

\paragraph{The Design Tension}
Our findings on the Textual Grounding (TG) adaptation reveal a critical design tension in parallelized web agents. The architecture's choice to process webpage sections in isolated batches creates a lack of global state, which interacts negatively with structured-input formats like TG. Without inter-batch communication, each batch independently selects its most plausible local action, unable to determine if the true target has already been found elsewhere. This architectural design consistently leads to an over-generation of viable candidate actions. The final `Action Selection' stage is therefore burdened not with identifying a correct action, but with disambiguating it from a flood of plausible but unnecessary alternatives, which directly results in lower accuracy. This demonstrates that efficiency gains from parallelization can be nullified if the agent lacks a mechanism to maintain a coherent, global understanding of the task.

\paragraph{Rethinking Benchmarking}
Our analysis, particularly the error classification and the performance boost from mined candidates, compellingly shows that a significant portion of measured "errors" are actually reasonable alternative solutions not captured by the single-ground-truth paradigm of benchmarks like Mind2Web. This has profound implications for the field. It calls for an urgent shift towards more flexible evaluation protocols that can accommodate multiple valid action paths. The continued reliance on rigid benchmarks not only inaccurately penalizes sophisticated models but also steers research away from solving real-world ambiguity. We echo the call for more dynamic, live-web evaluation environments \cite{zhou2023webarena} that can provide a more faithful assessment of an agent's true reasoning capabilities.

\section{Conclusion}

This study demonstrates the utility of modular evaluation for understanding and improving multi-step web agents. Using web navigation as a case study, we adapted the SeeAct framework to trace information flow across stages—from perception to action selection—by introducing targeted input modifications, refined prompting strategies, and a novel LLM-based Action Selection module. Our modular framework supports fine-grained evaluation through both LLM-based and algorithmic metrics.

Experiments across six models on the Mind2Web benchmark not only improved SeeAct's baseline performance but also uncovered key design insights for future web agents:
\begin{itemize}[noitemsep, topsep=1pt, leftmargin=*]
    \item \textbf{Section-aware reasoning:} Incorporating global page layout and structural context can aid batch-wise perception and decision-making.
    \item \textbf{Visual–semantic grounding:} Tightening the connection between screenshot regions and HTML markup is crucial for robust grounding.
    \item \textbf{Flexible supervision:} Supporting multiple valid actions, rather than relying on a single ground truth, better reflects the ambiguity and flexibility of real-world web tasks.
\end{itemize}

We hope this evaluation method encourages more interpretable, diagnostic evaluation for complex decision-making agents beyond the web setting.

\section{Limitations}

Our study demonstrates the value of modular evaluation through a case study on web navigation, but it is limited by its focus on a single framework (SeeAct) and benchmark. While modular evaluation is beneficial for NLP research, it may not apply to agents where intermediate steps lack direct alignment with ground truth. Additionally, although we evaluated six diverse V-LLMs, the inclusion of only one open-source model may underrepresent the performance variety in open-source SOTA models.

\bibliography{custom}

\appendix

\section{Full Prompt Example}
\label{sec:prompt_example}
Table \ref{tab:full_prompt_example} presents the adapted SeeAct prompting scheme used in our work.

\begin{table*}[htbp]
\label{tab:prompt}
\small
\centering
\begin{tabular}{l p{12cm}}

\hline
\textbf{System Role:} & Imagine that you are imitating humans doing web navigation for a task step by step. At each stage, you can see the webpage like humans by a screenshot and know the previous actions before the current step decided by yourself through recorded history. You need to decide on the first following action to take. You can click an element with the mouse, select an option, or type text with the keyboard.
(For your understanding, they are like the click(), select\_option() and type() functions in playwright respectively) One next step means one operation within the three. 
\\
\hline
\textbf{Action Prediction:} & {You are asked to complete the following task: \textit{\{task\}} \newline

Previous Actions: \textit{\{prev\_actions\}}
\newline

The screenshot below shows a section of a webpage. In this screenshot web elements of interest are outlined with red bounding boxes. Ensure to focus any actions on the highlighted elements. Follow the following guidance to think step by step before outlining the next action step at the current stage:\newline

(Current Webpage Identification)
Firstly, think about the purpose of this webpage section. Note that you are given section \textit{\{batch\_id\}/\{total\_batches\}} of the webpage.\newline

(Previous Action Analysis)
Secondly, combined with the screenshot, analyze each step of the previous action history and their intention one by one. Particularly, pay more attention to the last step, which may be more related to what you should do now as the next step.\newline

(Web Element Analysis)
The screenshot shows \textit{\{num\_candidates\}} web elements (such as links, buttons, and input fields) highlighted with red bounding boxes. Below is a textual description of the \textit{\{num\_candidates\}} highlighted elements:
\textit{\{choices\_simple\}} \newline

Select up to 5 elements that are most likely to be interacted with based on the current task and previous actions. For each of these 5 elements, describe its general function (e.g., "This date-picker allows the user to select a date for booking a flight.") and explain if interacting with this element is relevant to the task.\newline

(Next Action Based on Webpage and Analysis)
Then, based on your analysis, in conjunction with human web browsing habits and the logic of web design, decide on the following action.
Note that this section of the webpage may contain no viable element to interact with. In this case, you should issue a ``None'' action.
In case there is a viable action clearly outline which element in the webpage users will operate with as the first next target element, its detailed location, and the corresponding operation.\newline

To be successful, it is important to follow the following rules:  \newline
1. You should only issue a valid action given the current observation.   \newline
2. You should only issue one action at a time"""} \\
\hline
\textbf{Action Grounding:} & {(Reiteration)
First, reiterate your next target element, its detailed location, and the corresponding operation.\newline

(Multichoice Question)
Below is a multi-choice question where the choices correspond to the highlighted elements in the screenshot. The choices are sorted to correspond to their occurrence on the website (top-left to bottom-right). From the screenshot, find out where and what each one is on the webpage. Then, determine whether one matches your target element. Please examine the choices one by one. Choose the matching one. If multiple options match your answer, choose the most likely one by re-examining the screenshot, the choices, and your further reasoning.\newline

\textit{\{choices\}}\newline

(Final Answer)
Finally, conclude your answer using the format below. Ensure your answer is strictly adhering to the format provided below. Please do not leave any explanation in your answers of the final standardized format part, and this final part should be clear and certain. The element choice, action, and value should be in three separate lines.\newline

Format:\newline

ELEMENT: The uppercase letter of your choice.

ACTION: Choose an action from {{CLICK, TYPE, SELECT}}.

VALUE: Provide additional input based on ACTION.\newline

The VALUE means:
If ACTION == TYPE, specify the text to be typed.
If ACTION == SELECT, specify the option to be chosen.
If ACTION == CLICK, write "None".
"""} \\
\hline

\end{tabular}
\caption{Full Prompt Example}
\label{tab:full_prompt_example}
\end{table*}

\begin{figure*}[h]
  \includegraphics[width=\textwidth]{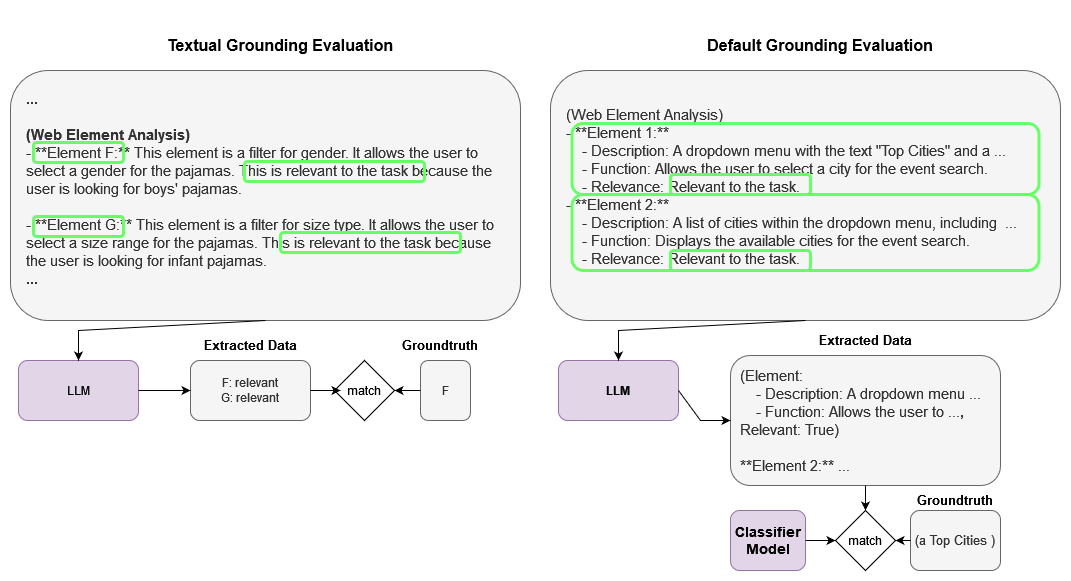} \hfill \centering
  \caption{Extraction \& Matching pipeline for Action Prediction evaluation}
  \label{fig:ap_eval_pipe}
\end{figure*}

\section{Evaluating Relevant Elements}
\label{appendix:evaluating_rel_elems}
In this section, we present the pipeline used to evaluate the Action Prediction stage, focusing on the creation of two key metrics: Relevant Element (RE Acc.) and Action Prediction Accuracy (AP Acc). These metrics assess whether the elements involved in  or the final action align with the ground truth. Additionally, for the intermediate reasoning stage, we determine whether each listed element is relevant to the task based on the LLM’s description.

Due to the non-deterministic structure of the Action Prediction output, we use an LLM (Gemini-1.5-Flash) to extract elements from the reasoning stage. This process generates a JSON object containing the extracted text for each element along with a Boolean value indicating its relevance, as determined by the description in the output. An overview of the pipeline and inputs is shown in Figure \ref{fig:ap_eval_pipe}.

We implemented two variants of the pipeline, depending on the presence of the Textual Grounding (TG) adaptation. TG simplifies the matching of extracted elements to the ground truth by providing a capital letter identifier, along with a compressed HTML representation, as input during the Action Prediction stage. The LLM detects this identifier (Figure \ref{fig:ap_eval_pipe}), allowing us to match it with the ground truth identifier.

\begin{table}[h]
\centering
\begin{tabular}{|c|c|}
\hline
\textbf{Metric} & \textbf{Value (\%)} \\
\hline
Accuracy & 86.18 \\
\hline
Precision & 79.28 \\
\hline
Recall & 82.59 \\
\hline
F1 Score & 80.09 \\
\hline
\end{tabular}
\caption{Results of Element matching via classifier}
\label{table:metrics_percent}
\end{table}

When TG is not enabled, the matching process becomes more complex, as the element descriptions rely solely on visual context, losing any predefined structure. In this case, we still use the LLM to extract listing elements but introduce a secondary classification stage to verify matches. This classifier, BGE-small, was trained using the SetFit framework on 800 manually labeled samples, with 100 additional samples used for evaluation. Each sample consists of an extracted listing element and its corresponding ground truth HTML representation. The 800 samples were drawn from Action Prediction outputs across four different LLMs (Gemini Flash/Pro, GPT 4o/mini) to ensure robustness to structural variations. Evaluation results are presented in Table \ref{table:metrics_percent}.

We also experimented with using LLMs directly for matching but found their performance to be suboptimal in both zero- and few-shot scenarios.

\section{Bounding-Box occlusion}
\label{appendix:bbox_occlusion}
\begin{figure}
    \centering
    \includegraphics[width=\columnwidth]{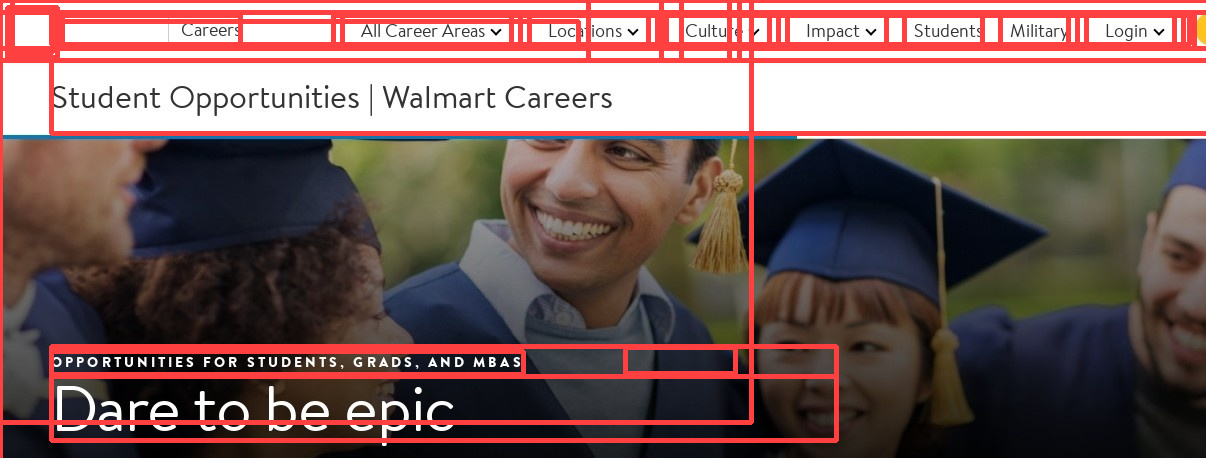}
    \caption{Bounding Box Occlusion Example (parameters chosen to favor occlusion)}
    \label{fig:bbox-occlusion example}
\end{figure}

The Visual Clues (VC, see Section \ref{subsubsec:input-modifi}) adaptation introduces red bounding boxes on webpage images to guide element selection during web navigation tasks.Below we illustrate the problems that can arise from creating bounding boxes based on the coordinates provided by the Mind2Web dataset \cite{deng2024mind2web}. An example screenshot of a webpage section is presented in Figure \ref{fig:bbox-occlusion example}. To provide a clear illustration of these issues, we adjusted pre-processing parameters to favor occlusion; in actual pipeline inference, these effects are typically less pronounced.

\paragraph{Excessive Nesting of Section Candidates} This issue occurs when the ranking model returns multiple hierarchically related elements for the same section. Consequently, multiple tightly packed bounding boxes may overlap, obstructing the content of neighboring web elements. Additionally, the bounding box of a parent container might intersect with surrounding elements, further complicating the visual representation.

\paragraph{Non-Visible Elements} Bounding boxes may also be created for elements that do not correspond to visible content in the image. This can potentially confuse the LLM's understanding of the drawn bounding boxes. Such elements include those without visible content (e.g., elements lacking text) or elements that are not currently displayed, such as dropdown menu items within a collapsed dropdown menu.

This section highlights the extra visual understanding required to fully benefit from Visual Clues. While addressing occlusion issues is beyond our scope, preliminary tests suggest that merging neighboring bounding boxes could mitigate them.

\section{Number of viable Actions}
\label{appendix: num_viable_actions}
Processing a webpage in multiple sections (batches) causes the Grounding stage to return an equal number of potential actions. These actions can be viable or include a "None" action, resulting from the LLM deciding not to act or a mismatch between the predicted action and the grounding candidate set. In the subsequent Action Selection stage, only viable actions are considered. Therefore, the number of viable actions determines the difficulty of the action selection process by setting the number of options to choose from.

Using the modular evaluation results, we calculated the average number of viable actions across models and adaptations (Table \ref{tab:num_viable_actions}).

\begin{table}[]
\begin{tabular}{lrrrr}
\toprule
      Model &   TG &   VC &  Def.\\
\midrule
  Gemini Flash & 4.32 & 2.99 & 2.68 \\
    Gemini Pro & 3.69 & 2.99 & 2.34 \\
    Claude Sonnet & 4.00 & 2.92 & 2.71 \\
    InternVL2 & 3.94 & 3.44 & 3.24 \\
     GPT-4o & 3.29 & 2.34 & 1.93 \\
GPT-4o-mini & 4.23 & 2.94 & 2.59 \\
\bottomrule
\end{tabular}
\caption{Average number of viable action per model and adaptation}
\label{tab:num_viable_actions}
\end{table} 

We find that TG consistently yields the most viable actions, followed by VC and then Def, suggesting that additional constraints during the Action Prediction stage lead to more actions returned. Note that TG was run with 5 sections while VC and Def were run with 4, based on optimal setups for each adaptation. Notably, GPT4-o produces the fewest viable actions across all three adaptations. InternVL2 results in the highest number in two out of three adaptations, closely matching the maximum in the third.

In our second analysis (Table \ref{tab:action_selection_performance}), we examined how the number of choices affects action selection accuracy. 

\begin{table}[]
\begin{tabular}{l l r r r}
\toprule
& \multicolumn{3}{c}{Number of Options} \\
\cmidrule(lr){2-4} 
Model & 2 & 3 & 4 \\
\midrule
  Gemini Flash     &   69.43 &  68.67 &  58.85 \\
  Gemini Pro      &   72.22 &  66.5 &  59.04 \\
  Claude          &   80.15 & 75.43 & 60.07 \\
  InternVL2       &  68.9 & 63.8 & 51.1 \\
  GPT-4o         &   75.2 &  68.7 &  52.8 \\
  GPT-4o-mini    &   72.92 &  58.39 &  53.9 \\
\bottomrule
\end{tabular}
\caption{LLM selection accuracy by number of options}
\label{tab:action_selection_performance}
\end{table}

We observed that performance declines as the number of actions increases. Claude consistently outperforms other models across all stages, while InternVL2 performs the worst. Since these metrics are based on the modular evaluation results, each model encounters varying numbers of selections with 2, 3, or 4 options depending on prior performance. Excluding Claude as an outlier, selection accuracy with two actions ranges from 68.9\% to 75.2\%. Indicating that even when faced with only two viable actions models face notable uncertainty.

\section{Model Specifications}
\label{appendix: model_specifications}
In Table \ref{tab:model_versions} we introduce the specific model versions. Each model was prompted with a temperature of 0.0 to ensure maximal reproducibility during experiments. 

\begin{table}[]
\begin{tabular}{ll}
\toprule
Model                & Version / Release date                                    \\
\midrule
Gemini 1.5 Flash     & 001 / May 2024                                            \\
Gemini 1.5 Pro       & 001 / May 2024                                            \\
Claude Sonnet        & 3.5/ 20.06.2024                                           \\
InternVL2-LLama3 76b & Quantized Version \\& from \href{https://huggingface.co/OpenGVLab/InternVL2-Llama3-76B-AWQ}{Huggingface}  \\
GPT4o                & 06.08.2024                                                \\
GPT4o-mini           & 18.07.2024   \\       
\bottomrule
\end{tabular}
\caption{Model Versions}
\label{tab:model_versions}
\end{table}

For the Open-Source model (InternVL2) a single A100 80GB GPU was utilized for inference.

\section{Web Element Ambiguity}
\label{appendix:ambiguity}
Below we give a brief overview of three scenarios in which the ambiguous choices influenced the web agents decision making:  
\begin{itemize} 
\item \textbf{Identical Compressed HTML}: Multiple buttons with the text "booking" cannot be distinguished by their representation, leading to ambiguity. This calls for including additional context of surrounding HTML elements to uniquely identify each button.
\item \textbf{Related HTML Elements}: The LLM may choose to interact with a parent element of the ground truth element. While the action would be viable when executed in the browser, it is considered incorrect by Mind2Web's evaluation criteria. 
\item \textbf{Sibling HTML Elements}: A checkbox filter may be accompanied by a link adjacent to it, where clicking either has the same effect. Mind2Web considers only one as the correct action. As SeeACT utilizes a compressed HTML representation (HTML repr.) that highlights salient features of web elements the representation of both may appear similar: (checkbox price range 50) vs. (a price 50)
\end{itemize}

\section{Selecting number of webpage sections}
\label{appendix_selecting_sections}
Our experiments stretch six models and three adaptations totaling 18 ablations. The main hyperparemter to set in the SeeAct framework \cite{zheng2024gpt} is the number of webpage sections (batches) into which the website is split for parallel processing. We decided on using four sections for Default (Def) and Visual Clues (VC) adaptaions as well as five for Textual Grounding (TG). We base this decision on selecting the optimal number of sections through a preliminary result where we ablated 4-5 sections using Gemini-1.5-Flash on the Website split (168 samples). Results of this preliminary study are provided in Table \ref{tab:num_batches_res}.
Based on the optimal LLM Select performance we chose the aforementiond number of batches.

\begin{table}[h]
\centering
\begin{tabular}{lrrr}
\toprule
\textbf{Metric} & \textbf{TG} & \textbf{VC} & \textbf{Def} \\
\midrule
\multicolumn{4}{c}{\textit{Num Batches = 4}} \\
\midrule
Grounding      & 46.15 & 42.01 & 46.7 \\
First Viable   & 22.02 & 19.6  & 25 \\
LLM Select     & 26.7  & \underline{29.7}  & \underline{36.3} \\
\midrule
\multicolumn{4}{c}{\textit{Num Batches = 5}} \\
\midrule
Grounding      & 54.51 & 50    & 49.41 \\
First Viable   & 18.45 & 22.19 & 22 \\
LLM Select     & \underline{32.7}  & 28.4 & 29.76 \\
\bottomrule
\end{tabular}
\caption{Results for varying number of batches on Website split (168) samples using Gemini-Flash}
\label{tab:num_batches_res}
\end{table}



\end{document}